%% file: eccv2022submissionCR.tex
\documentclass[runningheads]{llncs}
\usepackage{graphicx}

\usepackage{tikz}
\usepackage{comment}
\usepackage{amsmath,amssymb} 
\usepackage{color}

\usepackage[accsupp]{axessibility}  

\usepackage{cite}
\usepackage[flushleft]{threeparttable}
\usepackage{multirow}
\usepackage[labelfont=bf, font=footnotesize]{caption}
\usepackage{subcaption}
\DeclareMathOperator*{\argmax}{argmax}
\input{figure.tex}

\begin{document}
\pagestyle{headings}
\mainmatter
\def\ECCVSubNumber{7099}  

\title{Efficient Deep Visual and Inertial Odometry with Adaptive Visual Modality Selection} 

\titlerunning{Visual modality selective VIO}
%
\author{Mingyu Yang\thanks{Equally contributed first co-authors}\and
Yu Chen$^\star$ \and
Hun-Seok Kim}
\authorrunning{M. Yang et al.}
%
\institute{University of Michigan, Ann Arbor MI 48109, USA
\email{\{mingyuy,unchenyu,hunseok\}@umich.edu}\\
}

\maketitle

\begin{abstract}
In recent years, deep learning-based approaches for visual-inertial odometry (VIO) have shown remarkable performance outperforming traditional geometric methods. Yet, all existing methods use both the visual and inertial measurements for every pose estimation incurring potential computational redundancy. While visual data processing is much more expensive than that for the inertial measurement unit (IMU), it may not always contribute to improving the pose estimation accuracy. In this paper, we propose an adaptive deep-learning based VIO method that reduces computational redundancy by opportunistically disabling the visual modality. Specifically, we train a policy network that learns to deactivate the visual feature extractor on the fly based on the current motion state and IMU readings. A Gumbel-Softmax trick is adopted to train the policy network to make the decision process differentiable for end-to-end system training. The learned strategy is interpretable, and it shows scenario-dependent decision patterns for adaptive complexity reduction. Experiment results show that our method achieves a similar or even better performance than the full-modality baseline with up to $78.8\%$ computational complexity reduction for KITTI dataset evaluation. The code is available at \url{https://github.com/mingyuyng/Visual-Selective-VIO}.

\keywords{visual-inertial odometry, deep neural networks, long short-term memory, gumbel-softmax, adaptive learning}
\end{abstract}

\section{Introduction}

\figintro

Visual-inertial odometry (VIO) estimates the agent’s self-motion using information collected from cameras and inertial measurement unit (IMU) sensors. With its wide applications in navigation and autonomous driving, VIO became one of the most important problems in the field of robotics and computer vision. Compared with visual odometry (VO) methods \cite{cadena2016past, engel2017direct, mur2015orb, forster2014svo}, VIO systems \cite{qin2018vins, li2013high} incorporate additional IMU measurements and thus achieve more robust performance in texture-less environments and/or in extreme lightning conditions. However, classical VIO methods (not based on deep learning) rely heavily on manual interventions for system initialization and careful parameter tuning (e.g., number of features per frame, threshold of feature matching, and keyframe selection) for each test environment.  Besides, there are still significant challenges to deploying such systems with rapid calibration for fast-moving scenarios \cite{yang2018challenges}.

With the tremendous success of deep learning in various computer vision tasks \cite{krizhevsky2012imagenet, simonyan2014very, ren2015faster}, data-driven VIO methods \cite{clark2017vinet, chen2019selective, liu2021atvio, shamwell2018vision, han2019deepvio, almalioglu2019selfvio} have drawn significant attention to the community, and they achieve competitive performance in both accuracy and robustness in challenging scenarios. Compared with classical geometric-based methods, these learning-based VIO solutions extract better features using deep neural networks (DNN). In addition, they can learn a better fusion mechanism between visual and inertial features to filter out abnormal sensor data while training on large-scale datasets. However, such learning-based methods typically have significant overhead in computation and power consumption, which is not affordable to energy-constrained mobile platforms operating with low-cost, energy-efficient cameras and IMU sensors. 

Motivated by recent works that apply temporal adaptive inference to realize efficient action recognition \cite{wu2019adaframe, meng2020ar, meng2021adafuse, panda2021adamml} and fast text classification \cite{seo2017neural, campos2017skip, hansen2019neural}, we propose a new adaptive policy-based method to alleviate the high computational cost of deep learning-based VIO methods. The trained policy network opportunistically disables the visual (image) modality, as illustrated in Figure \ref{fig:fig_intro}, to reduce the computational overhead when the visual features do not contribute significantly to the overall pose estimation accuracy. We choose to dynamically disable the visual modality while keeping IMU always available because the image encoder is much more computationally demanding than the inertial encoder due to their modality dimension difference. Thus, skipping the image processing significantly reduces the overall computational complexity. Besides, visual information is not always necessary for an accurate pose estimation, especially when the motion state does not vary much over time. Thus, occasionally skipping unimportant image inputs does not necessarily degrade the odometry accuracy. For our method, the proposed policy uses sampling from a Bernoulli distribution parameterized by the output of a light-weight policy network. We adopt the Gumbel-Softmax trick \cite{jang2016categorical} to make the decision process differentiable. The model is trained to strike a balance between accuracy and efficiency with a joint loss. Our experiments demonstrate that our method significantly reduces computation (up to 78.8\%) without compromising VIO accuracy. Thus, the proposed framework is suitable for mobile platforms with limited computation resources and energy budgets. Also, our method is modal-agnostic and can be applied to any visual and inertial encoders with different structures. Moreover, the learned policy is interpretable and yields scenario-dependent decision patterns in various test sequences.

Overall, our contributions are summarized as follows: 

\begin{itemize}
\item We propose a novel method that adaptively disables the visual modality on the fly for efficient deep learning-based VIO. To the best of our knowledge, we are the first to demonstrate such a system reducing the complexity and energy consumption of deep learning-based VIO.   

\item A novel policy network is jointly trained with a pose estimation network to learn a visual modality selection strategy to enable or disable a visual feature extractor based on the motion state and IMU measurements. We adopt a Gumbel-Softmax trick to make the end-to-end system differentiable. 

\item The proposed method is tested extensively on the KITTI Odometry dataset. Experiments show that our approach achieves up to $78.8\%$ computation reduction without noticeable performance degradation. Furthermore, we show that the learned policy exhibits an interpretable behavior that depends on motion states and patterns.
\end{itemize}

\section{Related Works}

\subsection{Visual-inertial odometry}

Visual odometry (VO) is a process to estimate ego-motion from sequential camera images \cite{nister2004visual}, and it is extended to visual inertial odometry (VIO) including an IMU as an additional input. The datapath of conventional schemes typically consists of the following steps: feature detection, feature matching and tracking, motion estimation, and local optimization \cite{scaramuzza2011visual}. The VO/VIO system can be integrated into a simultaneous localization and mapping (SLAM) system \cite{mur2015orb,mur2017orb, qin2018vins} by performing additional steps of 3D environment mapping, global optimization, and loop closure. The performance of conventional VIO/SLAM systems is largely affected by visual feature matching and tracking accuracy, and the sensor fusion strategy. Hence, identifying superior handcrafted feature descriptors \cite{lowe1999object, rublee2011orb}, adaptive filtering \cite{li2013high} or nonlinear optimization \cite{leutenegger2015keyframe, hong2017visual} based sensor fusion schemes are key challenges of such methods.

In recent years, deep learning-based methods have achieved remarkable successes on various computer vision applications, including VIO. VINet \cite{clark2017vinet} is the first end-to-end trainable deep learning-based VIO where a DNN learns pose regression from the sequence of images and IMU measurements in a supervised manner. A long short-term memory (LSTM) network is introduced in VINet to model the temporal motion correlation. Later, Chen et al. \cite{chen2019selective} propose two different masking techniques that selectively fuse the visual and inertial features. ATVIO \cite{liu2021atvio} introduces an attention-based fusion function and uses an adaptive loss for pose regression. Some recent works also propose to learn the 6-DoF ego-motion through a self-supervised learning framework that does not require ground-truth annotations during training. Shamwell et al. \cite{shamwell2018vision} introduce VIOLearner that estimates the poses through a view-synthesis approach with multi-level error correction. DeepVIO \cite{han2019deepvio} improves VIO poses by additional self-supervision of optical flow, and similarly Almalioglu et al.\cite{almalioglu2019selfvio} demonstrate a self-supervised VIO based on depth estimation \cite{zhou2017unsupervised, godard2019digging}.

Whereas these prior works always rely on both the visual and inertial modality for each pose estimation, we propose a new framework to save the computation and power consumption overhead by opportunistically disabling the visual modality based on a learned strategy.

\subsection{Adaptive inference}
An adaptive inference scheme dynamically allocates computing resources based on each task input instance to minimize the redundant computation for relatively `easy’ task inputs. Several techniques for adaptive inference have been proposed including early exiting \cite{huang2017multi, teerapittayanon2016branchynet, bolukbasi2017adaptive}, layer skipping \cite{graves2016adaptive, wang2018skipnet, veit2018convolutional}, and dynamic channel pruning \cite{hua2019channel, yuan2020s2dnas, yang2021deep}. Recently, the idea of adaptive inference has been extend to sequential data (e.g., text \cite{seo2017neural, campos2017skip, hansen2019neural} and videos \cite{wu2019adaframe, meng2020ar, meng2021adafuse, panda2021adamml}) that are processed by recurrent neural networks (RNNs). Our technique is closely related to adaptive video recognition first proposed in \cite{wu2019adaframe}, which introduces a memory-augmented LSTM to select only the relevant frames for efficient action recognition by training with a policy gradient method. Similarly, AR-Net \cite{meng2020ar} learns a policy that dynamically selects more relevant image frames and also adjusts their resolutions. The training in AR-Net is simplified using the Gumbel-Softmax trick. Later, this idea was extended to adaptively selecting a proper modality \cite{panda2021adamml} or patches \cite{wang2021adaptive}. 
Our approach is motivated by these prior works to apply a similar framework to adaptive computation on deep learn-based VIO for the first time. We formulate it as a discrete-time pose regression problem that produces a pose estimation for every time step.

\section{Method}

\figmethod

The inputs for VIO are the monocular video frames $\{V_i\}_{i=1}^N$, IMU measurements $\{I_i\}_{i=1}^{Nl}$ captured with a sampling frequency $l$ times higher than the video frame rate, and the initial camera pose $P_1$. The goal of VIO is to estimate the camera poses $\{P_i\}_{i=2}^N$ for the entire path where $V_i \in\mathbb{R}^{3\times H \times W}$, $I_i \in \mathbb{R}^{6}$, and $P_i \in \textbf{SE}(3)$. One typical way to perform VIO is to estimate the 6-DoF relative pose $T_{t\to t+1}$ that satisfies $P_tT_{t\to t+1}=P_{t+1}$ using two consecutive images $V_{t\to t+1} = \{V_t, V_{t+1}\}$ and a set of IMU measurements $I_{t\to t+1} = \{I_{tl}, \dots, I_{(t+1)l}\}$ for the time index $t=1, 2, \dots, N-1$. The relative pose $T_{t\to t+1}$ can be further decomposed into a rotational vector $\phi_t\in \mathbb{R}^3$ containing Eular angles and a translational vector $v_t\in\mathbb{R}^3$. Our method learns a selection strategy that opportunistically skips the visual information $V_{t\to t+1}$ to reduce the computational complexity while maintaining the relative pose estimation accuracy.

\subsection{End-to-end neural visual-inertial odometry} \label{background}
End-to-end neural VIO methods \cite{clark2017vinet, chen2019selective, liu2021atvio} consist of a visual feature encoder $E_{visual}$ and an inertial feature encoder $E_{inertial}$ that extracts learned features from the input images and IMU measurements as follows: 
\begin{equation}
    x_t^v=E_{visual}(V_{t\to t+1}), \;\;\; x_t^i=E_{inertial}(I_{t\to t+1}).
\end{equation}
Typically, the visual feature encoder is much larger than the inertial feature encoder as the image dimension is much larger than that of the IMU measurement. 

Visual feature $x_t^v$ and inertial feature $x_t^i$ are combined as $z_t$ through concatenation \cite{clark2017vinet} or attention modules \cite{chen2019selective, liu2021atvio}. For accurate pose estimation, estimated motions and states of previous frames are used together with the newly extracted features of the current frame. Because of this temporally sequential nature of the problem, an RNN is typically employed to learn the  correlation within the sequence of motions. The RNN employes fully connected layers as the last step for the final 6-DoF pose regression as in: 
\begin{equation}
    (h_t, \hat{\phi}_t, \hat{v}_t) = \text{RNN}(z_t, h_{t-1}),
    \label{equ:regression}
\end{equation}
where $h_{t-1}$ and $h_t$ are the hidden latent vectors of the RNN at time $t$ and $t-1$. $\hat{\phi}_t$ and $\hat{v}_t$ denotes the estimated rotational vector and translational vector, respectively.

\subsection{Deep VIO with visual modality selection} \label{policy}

The overview of our proposed method is illustrated in Figure \ref{fig:fig_method}. As an adaptive method, we aim to learn a binary decision $d_t$ to determine whether the visual modality is not necessary and can be disabled without significant pose estimation accuracy degradation. We introduce a decision module where the decision $d_t$ is sampled from a Bernoulli distribution whose probability $p_t$ is generated by a light-weight policy network $\Phi$. The policy network takes the current IMU features $x_t^i$ and the last hidden latent vector $h_{t-1}$ that contains the history information as the input. Thus, we have
\begin{equation}
   p_t = \Phi(h_{t-1}, x_i^t),
\end{equation}
where $p_t \in \mathbb{R}^2$ denotes the probability of the Bernoulli distribution. To make the system end-to-end trainable, we sample the binary decision $d_t\in \{0, 1\}$ via the Gumbel-Softmax operation, 
\begin{equation}
   d_t \sim \text{GUMBEL}(p_t).
\end{equation}

The detail of training with Gumbel-Softmax is discussed in the next section. 
When $d_t=1$, visual features are enabled and the combined feature is obtained by concatenation of visual features and inertial features. On the other hand, when $d_t=0$, visual features are disabled thus we apply zero padding to replace visual features to keep the same input dimension for the following RNN. This can be expressed as:
\begin{equation}
    z_t = 
    \begin{cases}
        x_t^v \oplus x_t^i & \text{if $d_t = 1$ } \\
        \mathbf{0} \oplus x_t^i & \text{otherwise } \\
    \end{cases},
\end{equation}
where $\oplus$ denotes the concatenation operation. The combined feature $z_t$ is then fed to the RNN that produces the estimated pose outputs ($\hat{\phi}_t$ and $\hat{v}_t$) via regression as in equation (\ref{equ:regression}). In this paper, we adopt a two-layer LSTM for the pose estimation RNN.

\subsection{Training with Gumbel-Softmax} \label{gumbel}

Sampled $d_t$ that follows a Bernoulli distribution is discrete in nature and it makes the network non-differentiable. Thus, it is not trivial to train the policy network through back-propagation. One common choice is to use a score function estimator (e.g., REINFORCE\cite{williams1992simple,glynn1990likelihood}) to estimate the gradient through the `log-derivative trick'. However, that approach often has issues with slow convergence and high variance \cite{wu2018blockdrop} for many applications. As an alternative, we adopt the Gumbel-Softmax scheme \cite{jang2016categorical} to resolve non-differentiability by sampling from a corresponding Gumbel-Softmax distribution, which is essentially a reparametrization trick for categorical distributions \cite{rezende2014stochastic, kingma2013auto,mohamed2020monte}. Though reparameterization tricks may be less general than score function estimators, they usually exhibit several advantages such as lower variance and easier implementation.   

Consider a categorical distribution where the probability for the $k_{th}$ category is $p_k$ for $k = 1, ..., K$. Then, following the Gumbel-Max trick \cite{jang2016categorical}, a discrete sample $\hat{P}$ that follows the target distribution can be drawn by:
\begin{equation}
     \hat{P} = \argmax_k (\log p_k + g_k), \;\;\; k \in [1, 2, ..., K],
    \label{eq:discrete}
\end{equation}
where $g_k=-\log(-\log U_k)$ is a standard Gumbel distribution with a random variable $U_k$ sampled from a uniform distribution $U(0,1)$. Later, the softmax function is applied to relax the argmax operation to obtain a real-valued vector $\tilde{P}\in \mathbb{R}^K$ by a differentiable function as in
\begin{equation}
     \tilde{P}_k = \frac{\exp((\log p_k + g_k)/\tau)}{\sum_{j=1}^{K}\exp((\log p_j + g_j)/\tau)}, \;\;\; k = 1, 2, ..., K,
    \label{eq:continuous}
\end{equation}
where $\tau$ is a temperature parameter that controls the `discreteness' of $\tilde{P}$. When $\tau$ goes to infinity, $\tilde{P}$ tends to be a uniformly distributed vector, whereas $\tau\approx0$ makes $\tilde{P}$ close to a one-hot vector and indistinguishable from the discrete distribution. In our case, we only have two categories $K=2$ since we are dealing with a binary decision. During training, we sample the policy from the target Bernoulli distribution through (\ref{eq:discrete}) for the forward pass whereas the continuous relaxation (\ref{eq:continuous}) is used for the backward pass to approximate the gradient. 

\subsection{Loss function} \label{loss}
During training, we apply the mean squared error (MSE) loss to reduce the pose estimation error given by:
\begin{equation}
   \mathcal{L}_{pose} = \frac{1}{T-1} \sum_{t=1}^{T-1} ( \|\hat{v}_t-v_t\|_2^2 + \alpha \|\hat{\phi}_t-\phi_t\|_2^2),
\end{equation}
where $T$ is the sequence length of training.
$v_t$ and $\phi_t$ denote the ground-truth translational and rotational vectors. $\alpha$ is a weight to balance the translational loss and rotational loss. We set $\alpha=100$ as in the setting in prior supervised learning VO/VIO methods \cite{wang2017deepvo, clark2017vinet, xue2019beyond, chen2019selective, liu2021atvio}.

Besides, we apply an additional penalty factor $\lambda$ to every visual encoder usage to encourage disabling visual feature computations. During the training, we calculate the averaged penalty and denote it as the efficiency loss defined by: 
\begin{equation}
   \mathcal{L}_{eff} = \frac{1}{T-1} \sum_{t=1}^{T-1} \lambda d_t. 
\end{equation}

Finally, the end-to-end system is trained with the summation of the pose estimation loss and efficiency loss (\ref{eq:joint_loss}) to strike a balance between good accuracy and computational efficiency.
\begin{equation}
   \mathcal{L} = \mathcal{L}_{pose} + \mathcal{L}_{eff}
   \label{eq:joint_loss}
\end{equation}

\section{Experiments}

In this section, we conduct an ablation study on the penalty factor to compare the proposed adaptive scheme with the full-modality baseline  that always uses visual features. Results in this section will show that our proposed visual modality selection strategy can significantly reduces  computational overhead while maintaining a similar or better accuracy compared to the full-modality baseline.

\subsection{Experiment Setup}

\subsubsection{Dataset}

We evaluate our approach on KITTI Odometry dataset \cite{geiger2012we}, which is one of the most influential VO/VIO benchmarks. The KITTI Odometry dataset consists of 22 sequences of stereo videos, where Sequence \textit{00}-\textit{10} contain the ground-truth trajectory and Sequence \textit{11}-\textit{22} exclude the ground-truth for evaluation. Following the procedure in \cite{chen2019selective}, we train our model with Sequence \textit{00, 01, 02, 04, 06, 08, 09} and test with Sequence \textit{05, 07}, and \textit{10}. We exclude Sequence \textit{03} because of the lack of the raw IMU data. The images and ground-truth poses are recorded at 10 Hz and the IMU data is recorded at 100 Hz. The IMU data and images are not strictly synchronized. Thus, we interpolate the raw IMU data to time-synchronize it with the images and ground-truth poses. We use the monocular images from the left camera of KITTI Odometry dataset.

\subsubsection{Implementation Details}

During training, we resize all images to $512\times256$ and set the training subsequence length to 11. We have 11 IMU measurements between every two consecutive images and thus the dimension of the input IMU data is $6\times11$. For the visual encoder, we adopt the FlowNet-S network \cite{dosovitskiy2015flownet} (except for the last layer) pretrained on the FlyingChairs dataset \cite{dosovitskiy2015flownet} for optical flow estimation. A fully connected layer is attached at the end of the network to produce a visual feature of length 512. The inertial encoder contains three 1D-convolutional layers and a fully connected layer to generate the inertial feature of size 256. The pose estimation network contains a two-layer LSTM each with 1024 hidden units. At each time step, the hidden state of the last LSTM layer is passed through a two-layer multi-layer perceptron (MLP) to estimate the 6-DoF pose. The policy network is designed with a light-weight three-layer MLP. 

The training process consists of two stages: warm-up stage and joint-training stage. In the warm-up stage, we train the visual encoder, inertial encoder, and the pose estimation network for 40 epochs with a random policy where we have a $50\%$ chance to use the visual encoder at each time step. The learning rate is set to $5\times10^{-4}$ in this stage. Next, in the joint-training stage, we train all end-to-end components including the policy network for 40 epochs with a learning rate of $5\times10^{-5}$, and then decrease the learning rate to $1\times10^{-6}$ for additional 20 epochs. We set the initial temperature of Gumbel-Softmax to 5 and apply exponential decaying for each epoch with a factor of $-0.05$. We use Adam optimization with $\alpha=0.9$ and $\beta=0.999$, and the batch size is set to 16. 
During training, we always use the visual modality for the first frame to guarantee a qualified initial pose estimation. Similarly, during inference, we always enable the visual modality for the first pose estimation before we run the policy network without intervention for the rest of the path. Although the sequence length for training is set to 11, our method can run on any length of inputs for the inference. 

\subsubsection{Metric}
\tabelrmse
\tabeltrel
We calculate the root mean square error (RMSE) for the estimated translational vectors $\{\hat{v}_t\}_{t=1}^{N-1}$ and rotational vectors $\{\hat{\phi}_t\}_{t=1}^{N-1}$ of the entire path 
(i.e., $\sqrt{\frac{1}{N-1}\sum_{t=1}^{N-1}\|\hat{v}_t-v_t\|_2^2}$ and $\sqrt{\frac{1}{N-1}\sum_{t=1}^{N-1}\|\hat{\phi}_t-\phi_t\|_2^2}$). We also evaluate the relative translation/rotation error denoted by $t_{rel}$ and $r_{rel}$ for various subsequence path lengths such as 100, 200, ..., 800 meters as in \cite{geiger2012we}. To evaluate our policy network, we calculate the average usage rate of the visual modality and GFLOPS (giga floating-point operations per second).

\subsection{Main results}

\subsubsection{Ablation study on the penalty factor}

\tabelbaseline
\figpathi

We first test our method on KITTI using four different penalty factors: $1\times10^{-5}, 3\times10^{-5}, 5\times10^{-5} \text{and\;} 7\times10^{-5}$ to compare with the full modality baseline. For a fair comparison, we train the proposed and baseline full modality models with the same optimizer and common  hyperparameters including the number of epochs and learning rate. Since our method is non-deterministic with a random sampling process, we test our model with 10 different random seeds and show the average performance. In Table \ref{table:test_RMSE}, we present the average usage rate of the visual encoder, average GFLOPS, and average translational and rotational RMSE. It is observed that, as we gradually increase the penalty factor $\lambda$, both the usage of the visual encoder and system GFLOPS decrease as expected. In the meantime, as the visual encoder usage (and GFLOPS) drops, the translational RMSE becomes monotonically worse while the rotational RMSE does not show a monotonic behavior. This indicates that visual features do not necessarily always contribute to improving rotation estimation accuracy. 
A particular setting of $\lambda=3\times10^{-5}$ provides 78.8\% reduction in GFLOPS at the cost of a relatively small 14.3\% loss in translational RMSE while improving rotational RMSE by  23.6\%. We also conduct evaluations of $t_{rel}$ and $r_{rel}$ obtained from various subsequent path lengths and report the results in Table \ref{table:test_rel}. Similarly, our method achieves comparable accuracy to the fully modality baseline with $\lambda=1\times10^{-5}$ and achieves an even better result with $\lambda=3\times10^{-5}$ which results in 78.8\% lower GFLOPS. Very aggressive policy network settings at $\lambda=5\times10^{-5}$ and $\lambda=7\times10^{-5}$ experience mild performance degradation. Note that the standard deviation shown in the last column remains quite small, demonstrating the stability of our proposed method.  

\subsubsection{Comparison with sub-optimal selection strategies}

\figpathii

In this section, we compare our proposed method with two sub-optimal visual modality selection strategies: regular skipping and random sampling. For regular skipping, we train the model with a fixed selection pattern where the visual encoder is enabled every $n$ time indices. For random sampling, the visual modality is enabled with probability of $p$ for each time index. These two methods are trained with the same number of epochs, optimizer, learning rate decaying strategy, and the other hyperparameters as in our proposed method. For each penalty factor $\lambda$ applied to our method, we carefully choose a corresponding skipping rate parameter $n$ and probability $p$ such that all methods share a similar visual encoder usage. Table \ref{table:baseline} shows our method significantly outperforms those two sub-optimal policies especially for $t_{rel}$. We also plot the path trajectories based on estimated poses from all methods on Sequence \textit{05} in Figure \ref{fig:fig_path05} and Sequence \textit{07} in Figure \ref{fig:fig_path07} for comparison. The proposed method exhibits the most reliable trajectory among all evaluated policies. 

\subsubsection{Comparison with other VO/VIO baselines}

\tablestoa

Now, we compare our method with geometric (non-learning-based) methods such as ORB-SLAM2 \cite{mur2017orb} and VINS-Mono  \cite{qin2018vins} without loop closure, and also with state-of-the-art deep learning-based VO/VIO methods. Among those, deep learning-based self-supervised methods are \cite{godard2019digging, zou2020learning, shamwell2018vision, han2019deepvio}, and supervised learning methods are \cite{xue2018guided, xue2019beyond, liu2021atvio, chen2019selectfusion}. All self-supervised methods are trained on Sequence \textit{00-08} and tested on \textit{09-10}. Among supervised methods, \cite{xue2018guided} and \cite{xue2019beyond} are trained on Sequence \textit{00, 02, 08, 09}. The other methods use the same training set as ours. It can be seen that although our main goal is not necessarily maximizing the odometry accuracy, our method still achieves the best performance among all the supervised methods. Compared with the state-of-the-art self-supervised methods \cite{shamwell2018vision, han2019deepvio} (which are known to outperform supervised methods in general), our method achieves a competitive performance especially for Sequence \textit{05} and \textit{07} that belong to their training set. This demonstrates the robustness of our policy network and also the effectiveness of our network structure and training strategy.

\subsubsection{Interpretation of the learned policy}

\figinterp
\figusage

In Figure \ref{fig:fig_interpretation}, we present the visual interpretation of our learned policy evaluated on Sequence \textit{07} with $\lambda=5\times10^{-5}$. On the top left, we plot the local visual encoder usage with a color coding that represents the visual modality usage rate for a local window of 31 frames. Darker (lighter) colors represent lower (higher) usages. On the top right, we show the speed of the agent (a vehicle) at each time step where darker colors represent lower speed. An obvious correlation is observed between the visual modality usage and the speed which is also correlated with the turning angle. When the agent is moving slowly or making a turn, the policy network utilizes the visual modality less frequently. When the agent moves straight and fast, the visual encoder is activated more frequently. 

One explanation for this behavior is based on the inherent IMU's property that directly measures the angular velocity. Unlike visual feature based estimation, it is relatively easy to estimate the turning angle using IMU because it is obtained by simple first-order integration. However, estimating translation requires additional process with IMU measurements because it only measures the acceleration which is the second-order differential of translation, requiring a qualified initialization of the velocity. Thus, when the agent is moving fast, IMU-only estimation tends to make large translation errors and hence the policy network enables the visual modality more frequently to reduce the errors.

To provide more insights on the behavior of the policy network, we selected three short segments from the path (marked with red squares in Figure \ref{fig:fig_interpretation} top left) to show the decisions $d_t$ and corresponding probabilities to enable the visual modality ($p_t$, generated by the policy network) on the bottom of Figure \ref{fig:fig_interpretation}. We mark $d_t$ using blue pulses and $p_t$ using orange circles. The policy network exhibits a clear `integrate-and-fire' pattern where it immediately resets the probability to $\approx0$ after the visual encoder is activated, and it keeps increasing the probability until the visual modality is enabled again. The slope of increasing $p_t$  varies along the path. When the agent is making a sharp turn and moving slowly, $p_t$ tends to increase slower and thus the gaps between two visual modality usages are relatively large (segments \textit{a} and \textit{b}). When the agent moves fast and straight, $p_t$ surges much faster leading to smaller gaps to enable the visual encoder.

 To show the general trend, we also plot the visual modality usage versus angular velocity and speed over all test paths for two different $\lambda$'s in Figure \ref{fig:fig_usage}. We calculate the averaged visual encoder usage for the intervals of $[0,0.1)$, $[0.1, 0.2)$, $\dots$, $[0.6, 0.7)$  $\text{rad}/s$ for the angular velocity and the intervals of $[0, 2)$, $[2, 4)$, $\dots$, $[14, 16)$ $m/s$ for the speed. It is observed that the usage is closely related to the angular velocity and speed. The usage in general tends to decrease with higher angular velocity and lower speed, although there can be occasional spots where this observation does not necessarily hold since our method is stochastic in nature.

\section{Conclusion}

In this paper, we propose a novel deep learning-based VIO system that  reduces computation overhead and power consumption by opportunistically disabling the visual modality when the visual information is not critical to maintain the accuracy of pose estimation. To learn the selection strategy, we introduce a decision module to the neural VIO structure and end-to-end train it with the Gumbel-Softmax trick. Our experiments show that our approach provides up to 78.8\% computation reduction without obvious performance degradation. Our learned strategy significantly outperform simple sub-optimal strategies. Furthermore, the learned policy is interpretable and shows scenario-dependent adaptive behaviours. Our adaptive learning strategy is model-agnostic and can be easily adopted to other deep VIO systems.

\paragraph{\textbf{Acknowledgements}}
This work was supported in part by Meta Platforms, Inc. We also acknowledge Google LLC for providing GCP computing resources. 

\clearpage
%
%
\bibliographystyle{splncs04}
\bibliography{egbib}
\end{document}

%% file: figure.tex
\newcommand{\figintro}{
\begin{figure}[t]
    \centering
    \includegraphics[width=0.9\columnwidth]{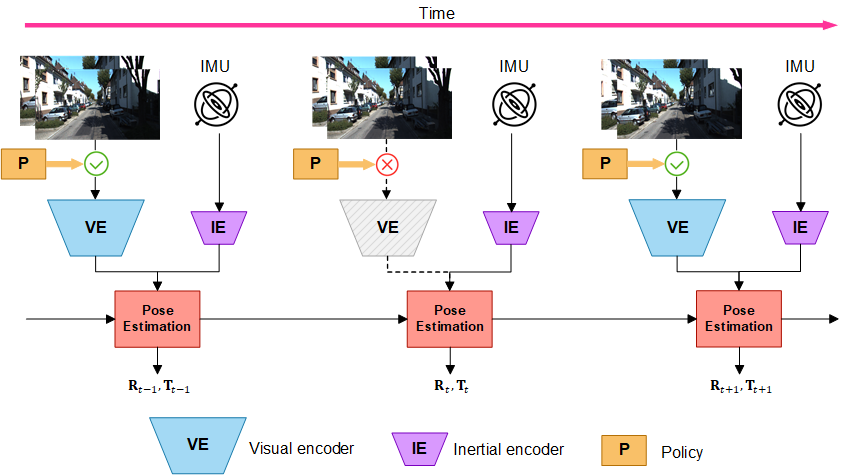}
    \caption{An overview of our approach. For deep learning-based VIO methods, the computational cost of the visual encoder is much higher than that of the inertial encoder due to the difference in data dimension. Thus, rather than using images for every pose estimation, our method learns a policy that controls the usage of the visual encoder to avoid unnecessary image processing while maintaining a reasonable accuracy. }
    \label{fig:fig_intro}
\end{figure}
}

\newcommand{\figmethod}{
\begin{figure}[t]
    \centering
    \includegraphics[width=0.9\columnwidth]{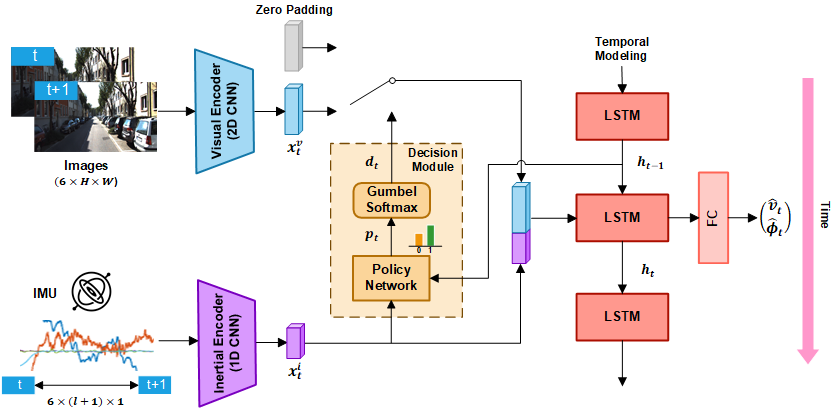}
    \caption{Illustration of the proposed framework. At each time step, the policy network takes the inertial features and the previous hidden state vector to decide whether to use the visual modality or not. Once the policy network decides to use the visual modality, the current image is passed through the visual encoder, and the corresponding visual features are fed to the pose regression LSTM together with the inertial features for pose estimation. Otherwise, the visual encoder is disabled to save computations and LSTM input is zero padded. The decisions are sampled from a Gumbel-Softmax distribution during training to make the system end-to-end differentiable. During inference, the decision is sampled from a Bernoulli distribution controlled by the policy network.}
    \label{fig:fig_method}
\end{figure}
}

\newcommand{\tabelrmse}{
    \begin{table}[t]
    \caption{Evaluation of the full-modality baseline and our proposed method with various penalty factors $\lambda$ on the KITTI dataset. Due to the stochastic nature of our policy, we test our model with 10 different random seeds and show the average performance.}
    \label{table:test_RMSE}
    \centering
    \scriptsize
    \begin{tabular}{c|c|c|c|c}
    \hline
    Method          & Trans. RMSE (m) & Rot. RMSE ($^{\circ}$) & Visual encoder usage     & GFLOPS   \\ \hline\hline
    Full Modality  & \textbf{0.0355}        & 0.0648 & 100\%     & 77.87      \\
    $\lambda = 1\times 10^{-5}$           & 0.0364        & 0.0505 & 62.89\%   & 49.04   \\
    $\lambda = 3\times 10^{-5}$         & 0.0406         & \textbf{0.0495}  & 21.02\%   & 16.51    \\
    $\lambda = 5\times 10^{-5}$          & 0.0477         & 0.0529    & 11.37\%   & 9.02      \\
    $\lambda = 7\times 10^{-5}$          & 0.0609  & 0.0592 & \textbf{6.85\%}   & \textbf{5.50} \\ \hline
    \end{tabular}
    \end{table}
}

\newcommand{\tabeltrel}{
    \begin{table}[t]
    \caption{The relative translational \& rotational error, and visual encoder usage of the baseline model and our proposed method with different penalty factors $\lambda$ on Sequence 05, 07, and 10. The results are averaged over 10 tests with different seeds. The last column also shows the standard deviation to quantify the stability.}
    \label{table:test_rel}
    \centering
    \begin{threeparttable}
    \tiny
    \begin{tabular}{c|ccc|ccc|ccc|ccc}
    \hline
    \multirow{2}{*}{Method}        & \multicolumn{3}{c|}{Seq. 05} & \multicolumn{3}{c|}{Seq. 07} & \multicolumn{3}{c|}{Seq. 10} & \multicolumn{3}{c}{Average}                     \\
           & $t_{rel}$       & $r_{rel}$  & Usage     & $t_{rel}$     & $r_{rel}$    & Usage  & $t_{rel}$       & $r_{rel}$   & Usage  & $t_{rel}$      & $r_{rel}$ &  Usage    \\ \hline\hline
    Full Modality &  2.61       & 1.06 & 100\%      & 1.83       & 1.35   & 100\%    & \textbf{3.11}       & 1.12  & 100\%     & 2.52       & 1.18  & 100\%  \\
    $\lambda = 1\times 10^{-5}$       & 2.15       & 0.78   & 60.30\%    & 2.25       & 1.19    & 63.35\%    & 3.30       & \textbf{0.94}   & 65.01\%    & $2.57\pm0.052$ & $0.97\pm0.018$ & 62.89\%    \\
    $\lambda = 3\times 10^{-5}$        & \textbf{2.01}       & \textbf{0.75}   & 20.60\%    & \textbf{1.79}       & \textbf{0.76}   & 19.79\%    & 3.41       & 1.08    & 22.68\%   & {$\mathbf{2.40\pm0.064}$} & $\mathbf{0.86\pm0.018}$    & 21.02\%  \\
    $\lambda = 5\times 10^{-5}$        & 2.71       & 1.03   & 11.34\%    & 2.22       & 1.14  & 10.57\%     & 3.59       & 1.20   & 12.20\%    & $2.84\pm0.102$ & $1.13\pm0.045$   & 11.37\%   \\
    $\lambda = 7\times 10^{-5}$        & 3.00       & 1.20   & 
    \textbf{6.83\%}    & 2.48       & 1.60   & \textbf{6.03\%}    & 3.67       & 1.57   & \textbf{7.68\%}    & $3.05\pm0.086$ & $1.46\pm0.046$   & \textbf{6.85\%} \\ \hline
    \end{tabular}

    \scriptsize
    \begin{tablenotes}
        \item[$\blacksquare$] $t_{rel}$ and $r_{rel}$ are the average translational error ($\%$) and average rotational error ($^\circ$/100m) obtained from various segment lengths of 100m -- 800m.
    \end{tablenotes}
    \end{threeparttable}
    \end{table}
}

\newcommand{\tabelbaseline}{
    \begin{table}[t]
    \centering
    \caption{Comparison with two sub-optimal policies (regular skipping and random sampling) that use a similar visual encoder usage.}
    \label{table:baseline}
    \scriptsize
    \begin{tabular}{c|c|c|c|c}
    \hline
    Method                            & params  & $t_{rel} (\%)$   & $r_{rel} (^\circ)$  & Visual encoder usage  \\ \hline\hline
    \multirow{2}{*}{Policy network}      & $\lambda=3\times 10^{-5}$    & $2.40\pm0.064$ & $0.86\pm0.018$ & 21.02\%   \\
                                  & $\lambda=5\times 10^{-5}$    & $2.84\pm0.102$ & $1.13\pm0.045$ & 11.37\% \\ \hline
    \multirow{2}{*}{Regular skipping} & $n = 5$   & 3.40  & 0.95  & 20\% \\
                                  & $n = 8$   & 5.39  & 2.15 & 12.5\% \\ \hline
                                  
    \multirow{2}{*}{Random policy}    & $p=0.2$   & $3.11\pm0.11$ & $1.16\pm0.073$ & 20.41\%  \\
                                  & $p=0.125$ & $4.42\pm0.239$ &  $1.2\pm0.027$     &  12.69\%   \\ \hline   
    \end{tabular}
    \end{table}
}

\newcommand{\figpathi}{
    \begin{figure*}[t]
        \centering
        \begin{subfigure}{0.49\linewidth}
        \centering\includegraphics[width=160pt]{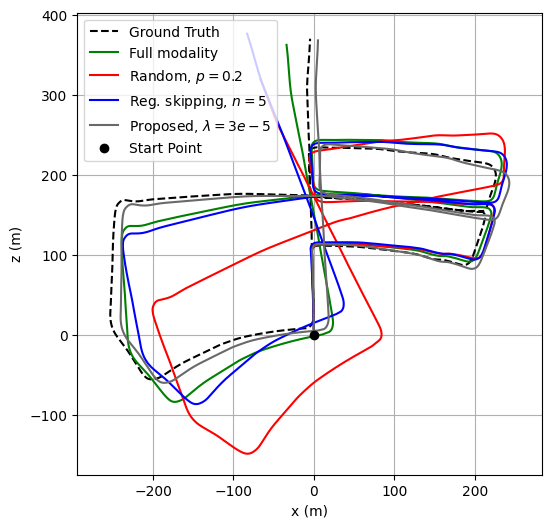}
        \end{subfigure}%
        \begin{subfigure}{0.49\linewidth}
        \centering\includegraphics[width=160pt]{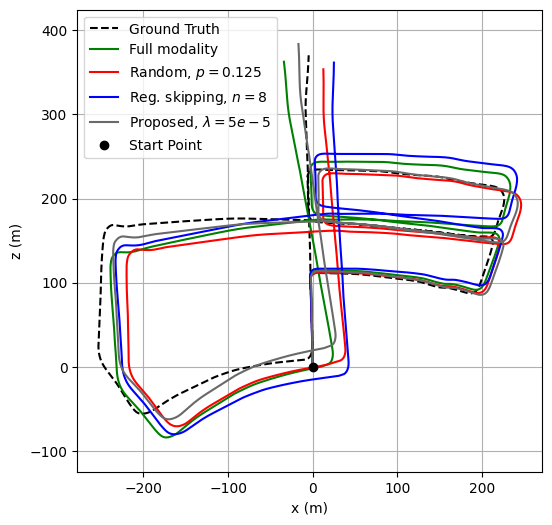}
        \end{subfigure}
    \caption{Trajectories of ground-truth, full modality baseline, random and regular skipping, and proposed method on KITTI Sequence \textit{05}. }
    \label{fig:fig_path05}
\end{figure*}
}

\newcommand{\figpathii}{
    \begin{figure*}[t]
        \centering
        \begin{subfigure}{0.49\linewidth}
        \centering\includegraphics[width=160pt]{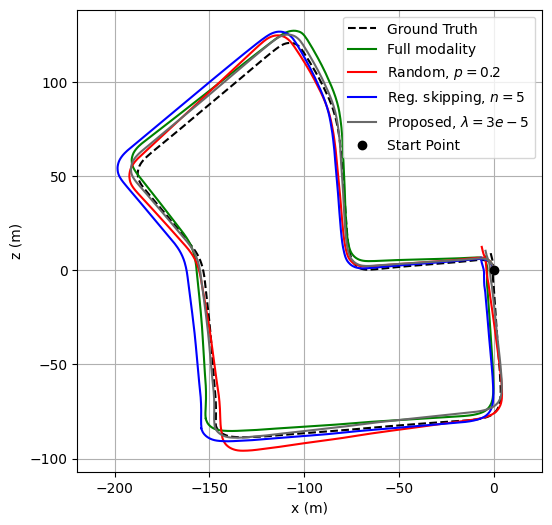}
        \end{subfigure}
        \begin{subfigure}{0.49\linewidth}
        \centering\includegraphics[width=160pt]{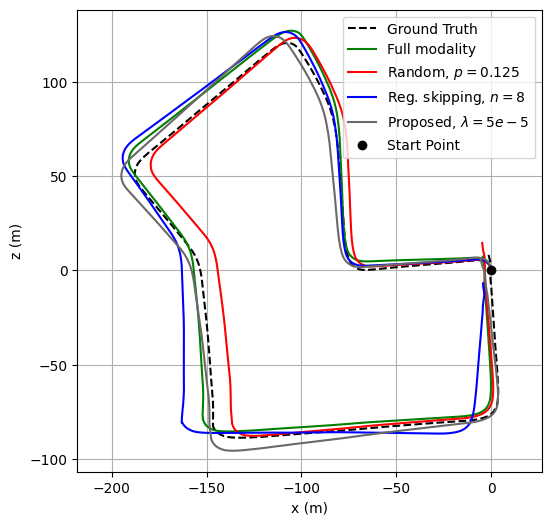}
        \end{subfigure}
    \caption{Trajectories of ground-truth, full modality baseline, random and regular skipping, and proposed method on KITTI Sequence \textit{07}. }
    \label{fig:fig_path07}
    \end{figure*}
}

\newcommand{\tablestoa}{

\begin{table}[t]
\tiny
\centering
\caption{Comparison with prior VO/VIO works in translational \& rotational error and image usage. The best performance in each block is marked in \textbf{bold}. Loop closure is excluded for ORB-SLAM2 and VINS-Mono.}

\label{table:headings}
\begin{threeparttable}
\begin{tabular}{cc|ccc|ccc|ccc}
\hline
& \multirow{2}{*}{Method}        & \multicolumn{3}{c|}{Seq. 05} & \multicolumn{3}{c|}{Seq. 07} & \multicolumn{3}{c}{Seq. 10}  \\
           &  & {\tiny $t_{rel}(\%)$}       & {\tiny$r_{rel}(^{\circ})$} & \tiny  Usage      & {\tiny $t_{rel}(\%)$}     & {\tiny$r_{rel}(^{\circ})$}   & \tiny Usage    & {\tiny $t_{rel}(\%)$}        & {\tiny$r_{rel}(^{\circ})$}  & \tiny Usage  \\ \hline\hline
\multirow{2}{*}{Geo.} & \tiny ORB-SLAM2\tnote{*} \cite{mur2017orb}      & \textbf{9.12}      & \textbf{0.2}   & 100\%    & 10.34       & \textbf{0.3}  & 100\%     & \textbf{4.04}       & \textbf{0.3}  & 100\% \\
&\tiny VINS-Mono\tnote{$\dagger$} \cite{qin2018vins}       & 11.6      & 1.26  & 100\%     & \textbf{10.0}       & 1.72   & 100\%    & 16.5       & 2.34   & 100\% \\ \hline
\multirow{4}{*}{\begin{tabular}{@{}c@{}c@{}} Self- \\ Sup. \end{tabular}} &\tiny Monodepth2 \tnote{*} \cite{godard2019digging}      & 4.66      & 1.7   & 100\%    & 4.58       & 2.6   & 100\%    & 7.73       & 3.4   & 100\% \\
&\tiny Zou et.al.\tnote{*} \cite{zou2020learning}      & \textbf{2.63}      & \textbf{0.5}   & 100\%    & 6.43       & 2.1   & 100\%    & 5.81      & 1.8   & 100\% \\
&\tiny VIOLearner\tnote{$\dagger$} \cite{shamwell2018vision}      & 3.00      & 1.40   & 100\%     & 3.60      & 2.06  & 100\%     & 2.04       & 1.37   & 100\% \\
&\tiny DeepVIO\tnote{$\dagger$} \cite{han2019deepvio}       & 2.86      & 2.32   & 100\%    & \textbf{2.71}       & \textbf{1.66}   & 100\%    & \textbf{0.85}       & \textbf{1.03}     & 100\% \\ \hline
\multirow{8}{*}{Sup.} &\tiny GFS-VO\tnote{*} \cite{xue2018guided}      & 3.27      & 1.6   & 100\%    & 3.37       & 2.2    & 100\%   & 6.32       & 2.3   & 100\%    \\
&\tiny BeyondTracking\tnote{*} \cite{xue2019beyond}      & 2.59      & 1.2    & 100\%  & 3.07       & 1.8    & 100\%   & 3.94       & 1.7   & 100\%    \\
&\tiny ATVIO\tnote{$\dagger$} \cite{liu2021atvio}      & 4.93      & 2.4   & 100\%    & 3.78       & 2.59   & 100\%    & 5.71       & 2.96    & 100\%   \\
&\tiny Soft Fusion\tnote{$\dagger$} \cite{chen2019selectfusion}      & 4.44      & 1.69  & 100\%     & 2.95       & 1.32   & 100\%   & 3.41       & 1.41   & 100\%     \\
&\tiny Hard Fusion\tnote{$\dagger$} \cite{chen2019selectfusion}      & 4.11      & 1.49   & 100\%    & 3.44       & 1.86  & 100\%     & \textbf{1.51}       & \textbf{0.91}   & 100\%    \\ \cline{2-11}
&\tiny (Ours) baseline\tnote{$\dagger$} &  2.61       & 1.06   & 100\%    & 1.83       & 1.35    & 100\%   & 3.11       & 1.12          & 100\%  \\
&\tiny (Ours) $\lambda=3\times 10^{-5}$\tnote{$\dagger$}\;          & \textbf{2.01}       & \textbf{0.75}   & 20.6\%    & \textbf{1.79}       & \textbf{0.76}    & 19.79\%   & 3.41       & 1.08     & 22.68\%   \\
&\tiny (Ours) $\lambda=5\times 10^{-5}$\tnote{$\dagger$}   \;       & 2.71       & 1.03   &  \textbf{11.34\%}    & 2.22       & 1.14    & \textbf{10.57\%}   & 3.59       & 1.20        & \textbf{12.2\%} \\ \hline 
\end{tabular}
*: Visual Odometry, $\dagger$: Visual-Inertial Odometry
\end{threeparttable}
\end{table}

}

\newcommand{\figinterp}{
\begin{figure}[t]
    \centering
    \includegraphics[width=0.9\columnwidth]{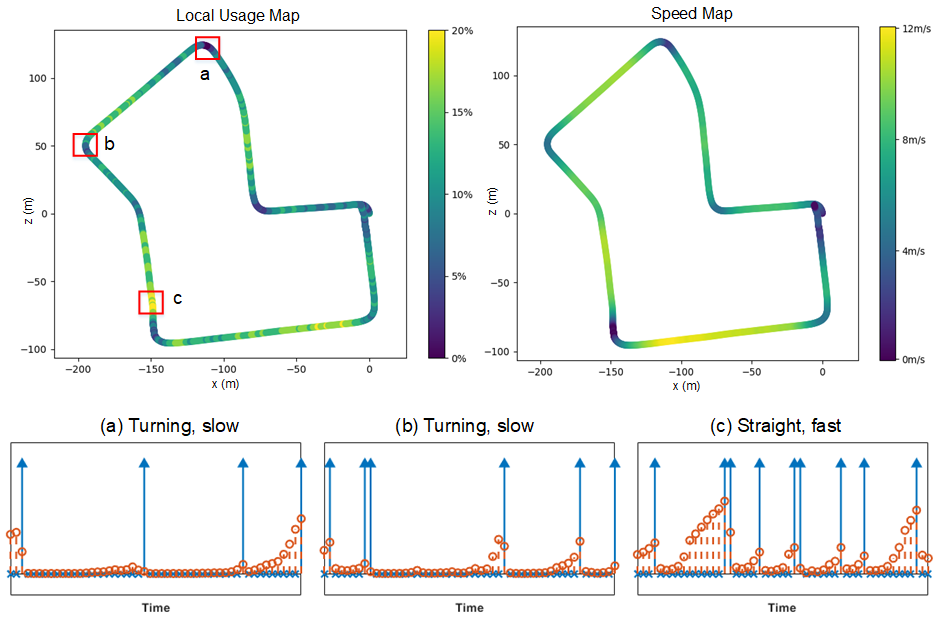}
    \caption{Visual interpretation of the learned policy on Sequence \textit{07} with $\lambda=5\times10^{-5}$. Top left is the usage map that shows the local usage rate at each time step calculated by averaging the activation rate of the visual encoder during a local window of 31 frames. The agent vehicle speed map is shown on the top right. We selected three short segments from the path to visualize the policy network's behavior by showing the decisions $d_t$ (blue pulses) and probabilities $p_t$ (orange circles) on the bottom for different scenarios. Seg. \textit{a} and \textit{b} show low speed with turning scenarios whereas Seg. \textit{c} is a high speed straight movement scenario. The policy network tends to activate the visual encoder more frequently when the agent is moving fast in straight, and decrease the usage of the visual encoder when the agent is moving slowly and making a turn. }
    \label{fig:fig_interpretation}
\end{figure}
}

\newcommand{\figusage}{
\begin{figure*}[t]
    \centering
    \begin{subfigure}{0.49\linewidth}
       \centering\includegraphics[width=160pt]{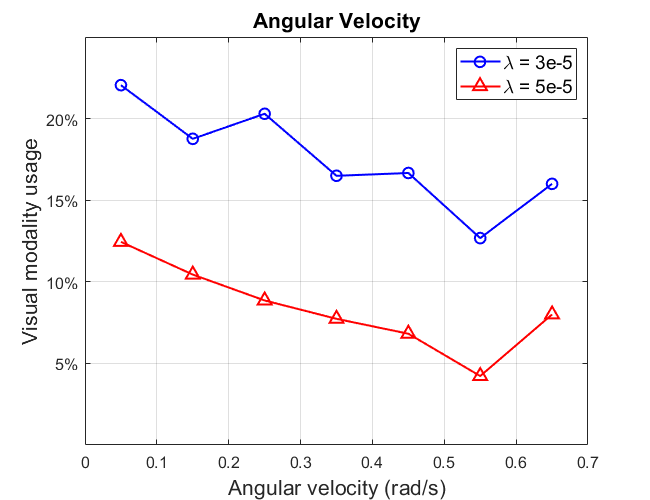}
    \end{subfigure}%
    \begin{subfigure}{0.49\linewidth}
       \centering\includegraphics[width=160pt]{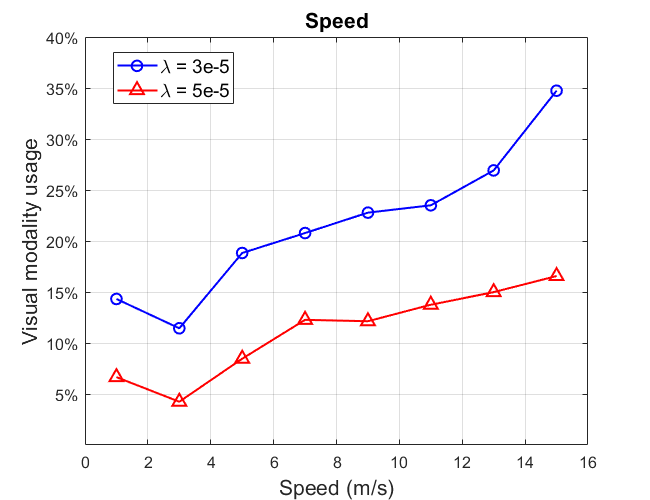}
    \end{subfigure}
\caption{The average usage rate of the visual modality for different angular velocities (left) and speeds (right) with two different penalty factors $\lambda$ over the entire test set (Sequence \textit{05, 07, 10}). The learned policy tends to use more images with lower angular velocity and higher speed.}
\label{fig:fig_usage}
\end{figure*}
}